\documentclass[10pt,twocolumn,letterpaper]{article}

\usepackage[pagenumbers]{wacv} 

\definecolor{wacvblue}{rgb}{0.21,0.49,0.74}
\usepackage[pagebackref,breaklinks,colorlinks,allcolors=wacvblue]{hyperref}
\usepackage{colortbl}
\usepackage{multirow}
\usepackage{xcolor}

\def\wacvPaperID{467} 
\def\confName{WACV}
\def\confYear{2026}

\title{Show Me: Unifying Instructional Image and Video Generation \\ with Diffusion Models}

\author{
Yujiang Pu$^{1}$ \quad Zhanbo Huang$^{1}$ \quad Vishnu Boddeti$^{1}$ \quad Yu Kong$^{1}$ \\
$^{1}$Michigan State University \\
{\tt\small \{puyujian, huang247, vishnu, yukong\}@msu.edu} \\
}

\begin{document}
\maketitle
\begin{abstract}
Generating visual instructions in a given context is essential for developing interactive world simulators. While prior works address this problem through either text-guided image manipulation or video prediction, these tasks are typically treated in isolation. This separation reveals a fundamental issue: image manipulation methods overlook how actions unfold over time, while video prediction models often ignore the intended outcomes.
To this end, we propose \textbf{ShowMe}, a unified framework that enables both tasks by selectively activating the spatial and temporal components of video diffusion models. In addition, we introduce structure and motion consistency rewards to improve structural fidelity and temporal coherence.
Notably, this unification brings dual benefits: the spatial knowledge gained through video pretraining enhances contextual consistency and realism in non-rigid image edits, while the instruction-guided manipulation stage equips the model with stronger goal-oriented reasoning for video prediction.
Experiments on diverse benchmarks demonstrate that our method outperforms expert models in both instructional image and video generation, highlighting the strength of video diffusion models as a unified action-object state transformer. Our code will be available at \textcolor{blue}{\small \url{https://yujiangpu20.github.io/showme/}}.

\end{abstract}

\section{Introduction}
\label{sec:intro}

Imagine you need to cook a steak or fold a shirt—what would you do? Most likely, you'd turn to a recipe or a YouTube video, reviewing it until you understand the steps. With the rise of large language models (LLMs)~\cite{brown2020language, touvron2023llama, achiam2023gpt}, you could even receive step-by-step instructions. However, while LLMs excel at generating detailed textual guidance, humans naturally interpret and learn through visual cues. YouTube videos, though visual, often present standardized procedures that may not align with a user’s specific context. This highlights the need for generating visual instructions that are tailored to the user’s situation, providing in-context, visually grounded guidance for complex tasks.




\begin{figure}[t]
    \centering
    \includegraphics[width=\linewidth]{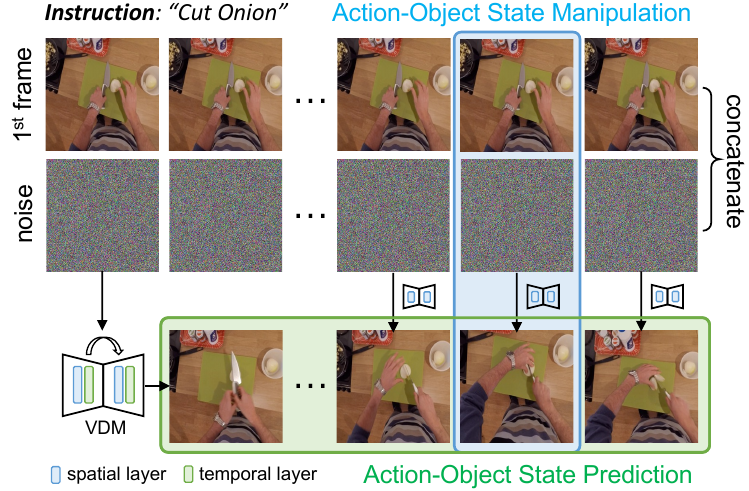}
    \caption{Video diffusion models inherently capture both spatial consistency and temporal dynamics, making them well-suited for unified action-object state transformations.}
    \label{fig:headfig}
\end{figure}

Despite its practical significance, visual instruction generation remains underexplored in the literature. LEGO~\cite{lai2024lego} is the first work to generate instructional images of human actions from an egocentric view. It utilizes LLaVA~\cite{liu2024visual} to produce enriched action descriptions, enabling diffusion models to achieve egocentric action manipulation. Concurrently, GenHowTo~\cite{souvcek2024genhowto} collects action frames along with action-state captions from instructional videos, enabling action and object state transformations by fine-tuning ControlNet~\cite{zhang2023adding}. Later, StackDiffusion~\cite{menon2024generating} generates illustrated instructions that meet user requirements by combining LLMs with text-to-image generation models. However, these efforts remain limited to creating ``static" instructions that solely depict action execution or final object states. 

To address this limitation, recent works have shifted toward instructional video generation. Seer~\cite{gu2023seer} formulates the task as video prediction conditioned on an initial frame and a natural language instruction. Meanwhile, UniSim~\cite{yang2023learning} builds a universal visual simulator by learning from diverse real-world and simulated data, enabling the prediction of visual outcomes based on both high-level instructions and low-level actions. AID~\cite{xing2024aid} further adapts the pretrained Stable Video Diffusion~\cite{blattmann2023stable} model using lightweight adapters and leverages LLaVA to produce step-by-step guidance as additional textual conditions. However, these studies primarily focus on overall generation quality, while overlooking the semantic alignment and spatiotemporal consistency required by specific instructions.

Despite sharing the goal of modeling action-object transformations, instructional image and video generation are typically studied in isolation. We argue that this separation introduces a fundamental limitation: image manipulation methods are blind to how actions unfold over time, while video prediction models often fail to capture the intended outcomes. This disconnect raises a natural question: \emph{Can a single model learn both to manipulate and to predict action-object transformations by sharing spatiotemporal priors?}

To this end, we propose \textbf{ShowMe}, a unified framework that enables both instructional image and video generation. 
Intuitively, we treat both tasks as two manifestations of action-object transformation, \emph{i.e.}, state manipulation and state prediction.
Our key insight is that pretrained video diffusion models (VDMs) inherently capture both spatial consistency and temporal dynamics. By decoupling the components and selectively tuning them via task-specific adapters, we demonstrate that a single VDM can be effectively repurposed to handle both tasks, as shown in \cref{fig:headfig}.

Specifically, we realize this framework through a two-stage tuning strategy. In the first stage, we disable the temporal modules of the VDM and fine-tune the spatial components using LoRA~\cite{hu2021lora} for action-object state manipulation, further guided by structural rewards derived from depth and edge priors. In the second stage, we reactivate the temporal layers and train a separate set of LoRA adapters for state prediction, guided by a motion consistency reward to encourage smooth and coherent dynamics. Notably, the pretrained LoRA from the first stage is retained and frozen during this phase, equipping the model with strong spatial reasoning to support faithful action execution. 
This unification benefits both tasks: video pretraining provides contextual coherence for more realistic action-object state manipulation, while instruction-guided manipulation enhances video prediction by promoting goal-oriented outcomes.

In summary, our contributions are as follows:
\begin{itemize}
    \item We propose ShowMe, a unified framework that repurposes VDMs as action-object state transformers, enabling both instructional image and video generation.
    \item We propose a two-stage tuning strategy with task-specific adapters while introducing structure and motion rewards to enhance spatial fidelity and temporal coherence.
    \item We demonstrate that ShowMe generalizes well across diverse image and video benchmarks, surpassing specialized baselines in both manipulation and prediction tasks.
\end{itemize}

\section{Related Work}
\label{sec:related}

\subsection{Diffusion-based Image Manipulation}

Image manipulation has advanced significantly with the advent of diffusion models, enabling more intuitive and controllable editing of visual content. SDEdit~\cite{meng2021sdedit} enables precise image synthesis by using stochastic differential equations to perturb and denoise images, leveraging the reverse process of diffusion models for guided manipulation. Later, Prompt-to-Prompt~\cite{hertz2022prompt} achieves training-free editing by controlling cross-attention layers in pre-trained diffusion models, enabling localized and fine-grained edits through modifications to the input text prompt. 
InstructPix2Pix~\cite{brooks2023instructpix2pix} further developed a large dataset of paired images and editing instructions, allowing models to accurately follow natural language commands for image editing without detailed prompts or masks. Despite these efforts, existing methods primarily focus on altering images' appearance, structure, or style, with little exploration of non-rigid edits related to human actions or object states. 
LEGO~\cite{lai2024lego} first proposes to create detailed action descriptions through visual instruction tuning, which also integrates image and text embeddings from LLaVA~\cite{liu2024visual} as additional conditions for egocentric action frame generation.
Concurrently, GenHowTo~\cite{souvcek2024genhowto} fine-tunes ControlNet~\cite{zhang2023adding} to learn action and object state transformations separately. A recent work, AURORA~\cite{krojer2025learning} explores the action and reasoning-centric image editing.

The work most related to ours is ShowHowTo~\cite{souvcek2025showhowto}, which also employs video diffusion models to generate contextually consistent step-by-step visual instructions. In contrast, we focus on state changes under single-step instructions and further extend this to video generation. Moreover, while ShowHowTo injects step-level instructions into frames through cross-attention, we adopt a spatio-temporal decoupling strategy to activate the editing capability and introduce reward tuning to improve structural fidelity. More quantitative and qualitative comparisons are presented in the following experiments.

\subsection{Video Diffusion Models}

Video diffusion models have garnered significant attention due to their potential in generating high-quality, temporally consistent frame sequences across various applications, including video editing~\cite{chai2023stablevideo,ceylan2023pix2video,feng2024ccedit,zhang2024towards}, interpolation~\cite{chen2023seine, lu2023vdt, danier2024ldmvfi,jain2024video}, and prediction~\cite{voleti2022mcvd,hoppe2022diffusion,yan2023feature,gu2023seer,zhang2024extdm}. VDM~\cite{ho2022video} is the first work extending 2D U-Net architectures to 3D space-time with factorized attention for video generation. To reduce training costs, subsequent works~\cite{singer2022make,ho2022imagen,blattmann2023align,guo2023animatediff,xing2024simda} have focused on extending text-to-image (T2I) models to text-to-video (T2V) generation by introducing additional temporal modules, showing impressive performance. Another research direction is (text-)image-to-video (I2V) generation~\cite{zhang2023i2vgen,blattmann2023stable,wang2024videocomposer,xing2025dynamicrafter,zeng2024make}, which aims to produce coherent video sequences based on conditional frames and text prompts. Additionally, some work has explored incorporating other control signals, such as pose~\cite{ma2024follow,hu2024animate}, structure~\cite{esser2023structure,xing2024make}, and motion~\cite{hu2023videocontrolnet,shi2024motion,liang2025movideo}, as conditions to generate personalized videos that align with user preferences. 

However, existing methods focus primarily on artistic creation or entertainment, leaving the conditional generation of human actions and object state changes underexplored. Seer~\cite{gu2023seer} takes an early step by extending Stable Diffusion~\cite{rombach2022high} for instruction-guided video prediction. UniSim~\cite{yang2023learning} builds a world simulator using a VDM to learn interactive control and decision-making from large-scale internet data. More recently, AID~\cite{xing2024aid} leverages LLaVA to generate detailed instructions and adapts SVD~\cite{blattmann2023stable} for video prediction.
In contrast, we curate instructional video datasets and introduce comprehensive benchmarks to evaluate both semantic alignment and spatiotemporal coherence, which are absent in these prior works.

\section{Method}

\begin{figure*}[t]
  \centering
   \includegraphics[width=\linewidth]{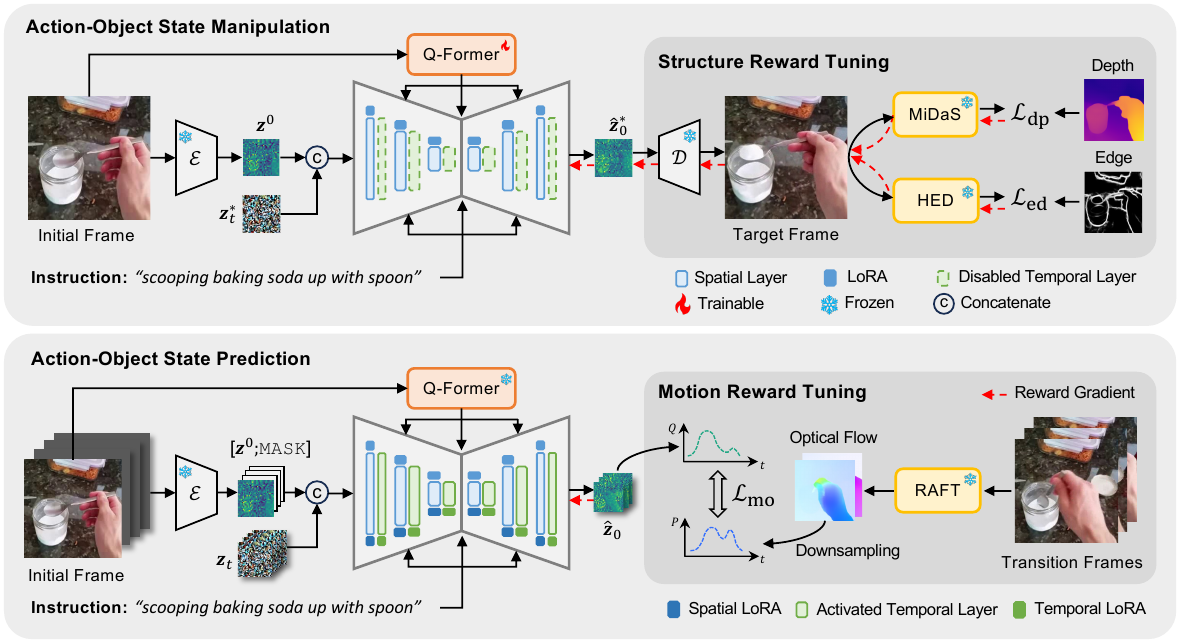}

   \caption{Illustration of the proposed \textbf{ShowMe} framework. For action-object state manipulation, we integrate LoRA into the Q-Former and spatial layers while disabling the temporal layers for model fine-tuning, followed by structure reward tuning to enhance depth and edge fidelity. For state prediction, we freeze all parameters and apply spatiotemporal LoRA for joint tuning, guided by motion reward to improve motion smoothness and consistency.}
   \label{fig:framework}
\end{figure*}

\noindent\textbf{Preliminaries.} Diffusion models~\cite{ho2020denoising, song2020score} are generative models that learn to produce data by simulating a gradual denoising process. Stable Diffusion~\cite{rombach2022high} enhances this process by leveraging a variational autoencoder (VAE)~\cite{kingma2013auto} to encode images as latent variables, denoted as $\mathbf{z}_0 = \mathcal{E}(\mathbf{x}_0)$. The forward process $q(\mathbf{z}_t|\mathbf{z}_0, t)$, which gradually adds noise to $\mathbf{z}_0$ over $T$ time steps, can be expressed as:
\begin{equation}
    \mathbf{z}_t = \sqrt{\Bar{\alpha}_t} \mathbf{z}_0 + \sqrt{1 - \Bar{\alpha}_t} \mathbf{\epsilon}, \mathbf{\epsilon} \sim \mathcal{N}(\mathbf{0}, \mathbf{I}),
    \label{eq:diffusion}
\end{equation}
 where $\Bar{\alpha}_t=\prod_{i=1}^t(1-\beta_t)$ controls the noise schedule. The reverse process is modeled as a learned Markov chain that approximates the posterior  $p(\mathbf{z}_{t-1} | \mathbf{z}_t)$ to denoise the latent representation. To this end, the model learns a denoising function $\epsilon_\theta$ to minimize the objective:
 \begin{equation}
    \mathcal{L}=\mathbb{E}_{\mathbf{z}_0, \mathbf{c}, \mathbf{\epsilon}\sim \mathcal{N}(\mathbf{0}, \mathbf{I}), t}\left[\left\|\mathbf{\epsilon}-\epsilon_\theta(\mathbf{z}_t, \mathbf{c}, t)\right\|^2_2\right],
    \label{eq:loss1}
\end{equation}
where $\mathbf{c}$ indicates the condition that achieves controllable generation. 
During sampling, the reverse process starts from random Gaussian noise and iteratively denoises to generate the final latent variable $\hat{\mathbf{z}}_0$, which is then projected back to pixel space using the decoder, \ie, $\hat{\mathbf{x}}_0=\mathcal{D}(\hat{\mathbf{z}}_0)$.

\noindent\textbf{Problem Definition.} 
In this work, we aim to tackle two closely related tasks: instructional image and video generation, both grounded in action-object transformations. Formally, given an initial image $\mathbf{x}^0 \in \mathbb{R}^{3 \times H \times W}$ and a textual instruction $\mathbf{c}$, our goal is to: (1) generate a target image $\hat{\mathbf{x}}^* \in \mathbb{R}^{3 \times H \times W}$ that reflects the execution of the instructed action with corresponding object state changes; and (2) generate a video $\hat{\mathbf{X}} = \{ \hat{\mathbf{x}}^0, \ldots, \hat{\mathbf{x}}^{L-1} \} \in \mathbb{R}^{L \times 3 \times H \times W}$ that illustrates the action’s progression over time. 

\noindent\textbf{Framework Overview.} 
We treat both tasks as two manifestations of action-object transformation, \emph{i.e.}, state manipulation and state prediction, and address them within a unified framework, as illustrated in \cref{fig:framework}.
For state manipulation, we disable the temporal modules of the pretrained VDM and apply LoRA to the Q-Former and spatial layers, unlocking inherent state reasoning capabilities. We further introduce structure consistency rewards to enhance spatial perception and grounding.
For state prediction, we activate the temporal modules and add separate LoRA adapters, followed by motion reward tuning to improve temporal coherence.
By selectively activating the spatial or temporal LoRAs, our method seamlessly switches between instructional image and video generation during inference.


\subsection{Action-Object State Manipulation}
Human actions and object states evolve over time, making video generation models well-suited as action-state transformers operating along the temporal dimension. Inspired by this, we repurpose a pretrained video diffusion model, DynamiCrafter~\cite{xing2025dynamicrafter}, for state manipulation.

The vanilla DynamiCrafter takes a reference image and a text prompt as conditions for video generation. To improve visual-text alignment, it extracts image embeddings using CLIP~\cite{radford2021learning} and integrates them via a Q-Former~\cite{li2023blip} projector.
To adapt the model for action-object state manipulation, we first flatten it into a 2D U-Net by disabling the temporal modules, which allows the model to focus solely on spatial reasoning while ignoring temporal dynamics.
We then insert LoRA~\cite{hu2021lora} modules into the attention layers of both the Q-Former and the spatial transformer for efficient fine-tuning, as shown in~\cref{fig:framework}.
Since the initial and final frames of an instructional video naturally represent an object’s starting and transformed states, respectively, we fine-tune the model with randomly sampled in-context video frame pairs $(\mathbf{x}^0, \mathbf{x}^*)$, supervised with the standard noise prediction objective defined in \cref{eq:loss1}.




Although the fine-tuned model can manipulate images based on the action instruction, we observe that it often struggles to preserve the appearance and structure of the manipulated content. 
We argue that this issue arises from the model’s limited ability to accurately perceive spatial relationships and locations within the 2D image context. 
To address this, we introduce a reward tuning strategy that incorporates two forms of structural information, \ie, depth and edge, to guide the generation process.
Existing methods~\cite{prabhudesai2023aligning, clark2023directly,xu2024imagereward} mainly introduce the differentiable reward during the reverse process, resulting in significant time and computational costs. In contrast, we adopt a one-step approximation that enables efficient target image supervision during training. Moreover, our rewards focus on structure preservation by aligning with ground-truth geometry, whereas DRaFT~\cite{clark2023directly} and AlignProp~\cite{prabhudesai2023aligning} primarily optimize for human preference alignment.


Specifically, we consider noisy latents from earlier diffusion steps $t$ ($t\leq\Gamma < T$) in the forward process as sampling targets. As noted in ~\cite{ho2020denoising, bansal2023universal}, such latents can be approximately denoised in a single step using the derivation from~\cref{eq:diffusion} and~\cref{eq:loss1}:
\begin{equation}
    \mathbf{z}_0^* \approx \hat{\mathbf{z}}_0^*=\frac{\mathbf{z}_t^* - \sqrt{1-\Bar{\alpha}_t}\epsilon_\theta(\mathbf{z}_t^*, \mathbf{c}, t)}{\sqrt{\Bar{\alpha}_t}},
     \label{eq:one-step}
\end{equation}
where $\hat{\mathbf{z}}_0^*$ is the estimated clean latent, which is then decoded into an RGB image via the VAE decoder $\mathcal{D}$. 
To introduce structural guidance, we leverage pretrained depth estimation~\cite{ranftl2020towards} and edge detection~\cite{xie2015holistically} models. As shown in \cref{fig:framework}, the decoded image is passed through a depth estimator and an edge detector. Both reward models are frozen during fine-tuning. We then compute structural rewards $\mathcal{L}_{\text{dp}}$ and $\mathcal{L}_{\text{ed}}$ by measuring the mean squared error between the predictions and the corresponding ground-truth maps. The overall reward objective is defined as:
\begin{equation}
    \mathcal{L}^*=\frac{1}{N} \sum_{i=1}^{N} \left\| \mathcal{R}_i(\mathcal{D}(\hat{\mathbf{z}}_0^*))-\mathcal{R}_i(\mathbf{x}^*)\right\|_1,
\end{equation}
where $\mathcal{R}_i$ refers to the $i$-th reward model (\eg, depth or edge). 
By jointly optimizing this reward loss with the noise prediction objective, our method encourages structural fidelity and spatial coherence in the generated images. Qualitative improvements are illustrated in \cref{fig:ssreward}.

\subsection{Action-Object State Prediction}
By adapting the video diffusion model for instructional image generation, we equip it with enhanced visual grounding and object-state reasoning capabilities. Building on this foundation, the next challenge is to generate temporally coherent visual transitions from a given initial state, \ie, to predict how an action unfolds over time.

The vanilla DynamiCrafter randomly samples video frames as conditions and concatenates them with repeated noisy latents to prevent positional shortcuts. In contrast, our task requires maintaining a fixed visual context to guide true prediction. Therefore, we fix the initial image at the starting position and apply zero padding as an unconditional mask, as shown in \cref{fig:framework}. 
We then activate the temporal modules and introduce a separate set of LoRA layers for spatiotemporal fine-tuning. Importantly, the pretrained LoRA modules from the manipulation stage are retained and frozen during this phase, providing strong spatial priors to support realistic state predictions.

\begin{figure}[t]
    \centering
    \includegraphics[width=\linewidth]{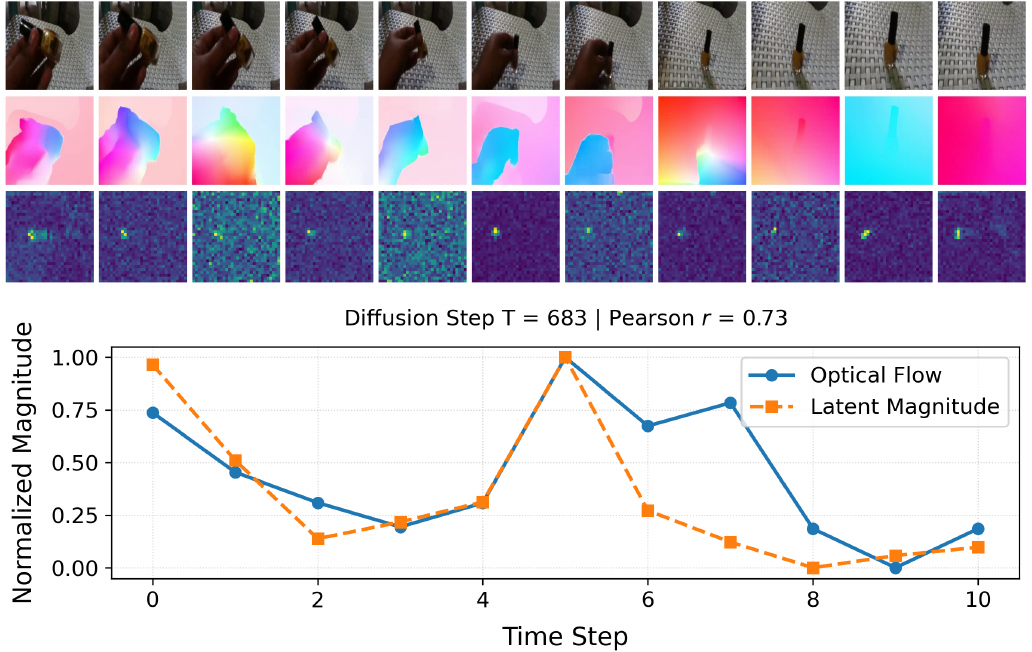}
    \caption{Motion consistency between flow and latent magnitude.}
    \vspace{-4mm}
    \label{fig:motion_consist}
\end{figure}

However, we observe that this unconditional mask can result in abrupt and unrealistic motion patterns. To address this, we introduce a motion consistency reward to facilitate spatiotemporal coherence and smoothness.
Intuitively, we can approximate clean latents via \cref{eq:one-step} and decode the full sequence back to pixel space for motion estimation. However, 
this inevitably incurs prohibitive memory and computational costs. A natural question arises: \emph{Can we apply motion rewards to the latent features without decoding them?} 
We further visualize the ground-truth optical flow in pixel space alongside the temporal magnitude of one-step ``denoised" latent features and observe that they exhibit strong temporal correlation, as shown in \cref{fig:motion_consist}. 
Based on this observation, we propose to align the distribution of temporal magnitudes of latent features with the optical flow to improve motion smoothness and consistency.

Notably, unlike structural rewards that rely on low-level noisy samples, we find that motion information remains present even in highly noisy latents, as observed in~\cite{burgert2025go}. Therefore, we perform one-step denoising on noisy latents from late diffusion steps using~\cref{eq:one-step}, and then compute the temporal magnitude of latent features as $\Delta\hat{\mathbf{z}}_{i}=\left\| \hat{\mathbf{z}}_{i+1}-\hat{\mathbf{z}}_{i}\right\|_2$. Meanwhile, we extract ground-truth optical flow using RAFT~\cite{teed2020raft} and downsample it to match the latent resolution, denoted as $\{\hat{\mathbf{f}}_i\}_{i=1}^{L-1}$. We treat both the latent motion and pixel-space flow magnitude as temporal distributions at each spatial location. Specifically, we normalize the latent magnitude sequence and the downsampled flow magnitude sequence over the temporal dimension:
\begin{align}
Q_i(h, w) &= \frac{\Delta\hat{\mathbf{z}}_i(h, w)}{\sum_{j=1}^{L-1} \Delta\hat{\mathbf{z}}_j(h, w)}, \\
P_i(h, w) &= \frac{\hat{\mathbf{f}}_i(h, w)}{\sum_{j=1}^{L-1} \hat{\mathbf{f}}_j(h, w)}.
\end{align}
We then compute the Kullback-Leibler (KL) divergence of the two temporal distributions to obtain the motion reward:
\begin{equation}
\mathcal{L}_{\text{mo}} = \mathbb{E}_{h, w} \left[ \mathrm{KL}(P \parallel Q) \right].
\end{equation}
This objective encourages the model to follow realistic motion patterns, promoting smoother and more coherent predictions. Moreover, it eliminates the need to decode latents back to pixel space, allowing the model to process entire sequences and better preserve motion continuity.





\section{Experiments}

\begin{table*}[t]
\centering
\resizebox{\textwidth}{!}{ 
\begin{tabular}{lcccccc|cccccc}
\toprule
\multirow{2}{*}{Method} & \multicolumn{6}{c|}{Something-Something V2} & \multicolumn{6}{c}{Epic-Kitchens 100} \\ 
\cmidrule(lr){2-7} \cmidrule(lr){8-13} 
& CLIP-I $\uparrow$  & CLIP-T $\uparrow$ & DINO-I $\uparrow$ & FID $\downarrow$  & PSNR $\uparrow$ & LPIPS $\downarrow$  & CLIP-I $\uparrow$ & CLIP-T $\uparrow$ & DINO-I $\uparrow$ & FID $\downarrow$ & PSNR $\uparrow$ & LPIPS $\downarrow$  \\ 
\midrule
ControlNet~\cite{zhang2023adding} 
& 0.8030 & 0.2870 & 0.5085 & 36.99 & 11.07 & 0.4590
& 0.8103 & 0.2532 & 0.5483 & 21.96 & 11.54 & 0.4866 
\\ 
SDEdit~\cite{meng2021sdedit}
& 0.7809 & \textbf{0.2923} & 0.4249 & 34.60 & \underline{13.30} & 0.4384
& 0.8115 & 0.2537 & 0.5373 & 16.00 & 12.65 & 0.4607 
\\ 
InstructPix2Pix~\cite{brooks2023instructpix2pix} 
& 0.8356 & \underline{0.2902} & 0.5827 & 31.93 & 12.46 & \underline{0.3868}
& 0.8395 & \underline{0.2562} & 0.6233 & 15.33 & 12.30 & 0.4358
\\ 
GenHowTo~\cite{souvcek2024genhowto} 
& 0.8062 & 0.2813 & 0.5735 & \underline{30.80} & \textbf{13.42} & 0.3921
& 0.8258 & 0.2528 & \underline{0.6529} & 15.70 & \textbf{12.82} & \underline{0.4293}
\\
AURORA~\cite{krojer2025learning} 
& 0.8349 & 0.2894 & 0.5815 & 32.46 & 12.26 & 0.3927
& 0.8423 & \textbf{0.2569} & 0.6326 & 15.19 & 12.41 & 0.4314
\\ 
ShowHowTo~\cite{souvcek2025showhowto} 
& \underline{0.8428} & 0.2871 & \underline{0.5875} & 31.32 & 11.23 & 0.4487
& \underline{0.8539} & 0.2551 & 0.6290 & \underline{14.18} & 12.54 & 0.4385
\\
\cellcolor{gray!25}ShowMe (Ours)
& \cellcolor{gray!25}\textbf{0.8561} & \cellcolor{gray!25}0.2860 & \cellcolor{gray!25}\textbf{0.6312} & \cellcolor{gray!25}\textbf{26.97} & \cellcolor{gray!25}12.84 & \cellcolor{gray!25}\textbf{0.3778}
& \cellcolor{gray!25}\textbf{0.8617} & \cellcolor{gray!25}0.2462 & \cellcolor{gray!25}\textbf{0.6607} & \cellcolor{gray!25}\textbf{12.96} & \cellcolor{gray!25}\underline{12.72} & \cellcolor{gray!25}\textbf{0.4232}
\\ 
\bottomrule
\end{tabular}
}
\caption{Performance of different image manipulation methods on instructional image generation. The compared methods are fully fine-tuned while our model only applies LoRA tuning. The best results are highlighted in \textbf{bold} and the second best is \underline{underlined}.}
\label{tab:ffp-sota}
\end{table*}

\begin{table*}[t]
\centering
\resizebox{\textwidth}{!}{ 
\begin{tabular}{lcccccc|cccccc}
\toprule
\multirow{2}{*}{Method} & \multicolumn{6}{c|}{Something-Something V2} & \multicolumn{6}{c}{Epic-Kitchens 100} \\ 
\cmidrule(lr){2-7} \cmidrule(lr){8-13} 
& FVD $\downarrow$ & FID $\downarrow$  & CLIP-I $\uparrow$ & ViCLIP $\uparrow$ & VQA $\uparrow$ & Motion $\uparrow$  & FVD $\downarrow$ & FID $\downarrow$ & CLIP-I $\uparrow$ & EgoVLP $\uparrow$  & VQA $\uparrow$ & Motion $\uparrow$ \\ 
\midrule

AnimateAnything~\cite{dai2023animateanything} 
& 263.48 & \underline{10.76} & \textbf{0.8752} & 0.1788 & 0.4704 &  10.19 
& 263.59 & \textbf{5.04} & \textbf{0.8796} & 0.3198 & 0.5629 & 22.48 
\\ 

Seer~\cite{gu2023seer} 
& 144.55 & 18.10 & 0.8324 & 0.1865 & 0.4601 & \underline{93.77} 
& 374.96 & 77.85 & 0.7491 & 0.2564 & 0.4821 & 91.03 
\\ 

ConsistI2V~\cite{ren2024consisti2v} 
& \underline{86.25} & \underline{10.76} & 0.8479 & \underline{0.1982} & 0.5093 & 90.30 
& \underline{90.45} & 10.20 & \underline{0.8638} & \underline{0.3535} & \underline{0.5753}  & 87.92 
\\ 

DynamiCrafter~\cite{xing2025dynamicrafter}  
& 108.78 & 15.58 & 0.8193 & 0.1932 & \underline{0.5108} & 90.77 
& 119.29 & 15.00 & 0.8366 & 0.3343 & 0.5526 & \textbf{97.67} 
\\ 

\cellcolor{gray!25}ShowMe (Ours) 
& \cellcolor{gray!25}\textbf{74.66} & \cellcolor{gray!25}\textbf{10.21} & \cellcolor{gray!25}\underline{0.8584} & \cellcolor{gray!25}\textbf{0.2012}& \cellcolor{gray!25}\textbf{0.5364} & \cellcolor{gray!25}\textbf{94.07} 
& \cellcolor{gray!25}\textbf{67.36}& \cellcolor{gray!25}\underline{9.01} & \cellcolor{gray!25}0.8637 &  \cellcolor{gray!25}\textbf{0.3537} & \cellcolor{gray!25}\textbf{0.5815} & \cellcolor{gray!25} \underline{96.72} 
\\ 
\bottomrule
\end{tabular}
}
\caption{Performance of different I2V methods on instructional video generation. The compared methods are fully fine-tuned while our model only applies LoRA tuning. The best results are highlighted in \textbf{bold} and the second best is \underline{underlined}.}
\label{tab:tvp}
\end{table*}



\noindent\textbf{Datasets.} We evaluate our method on two video datasets: Something-Something V2~\cite{goyal2017something} and Epic-Kitchens 100~\cite{damen2022rescaling}.
SSv2 is a large dataset focusing on human interactions with everyday objects, consisting of 220,847 short video clips across 174 action categories, such as “picking up a pen” or “folding a shirt.” We manually inspect the action labels and filter out static instructions like “holding” and “showing,” as well as videos depicting “pretending to do” actions, resulting in 112,321 training samples and 2,048 test samples. 
Epic100 captures daily kitchen activities from an egocentric perspective, comprising 100 hours of video footage with detailed narrations, including instructions such as “cut onion” and “stir vegetables.” Similarly, we filter out videos featuring highly dynamic scenes or noticeable scene transitions, obtaining 57,602 training samples and 8,236 test samples.
More details can be found in the supplementary material.

\noindent\textbf{Evaluation Metrics.} For instructional image generation, we employ six metrics to evaluate the model’s performance: CLIP-I~\cite{radford2021learning} and DINO-I~\cite{oquab2023dinov2} for contextual consistency; CLIP-T for semantic alignment; Fréchet Inception Distance (FID)~\cite{heusel2017gans} for realism; Peak Signal-to-Noise Ratio (PSNR) and Learned Perceptual Image Patch Similarity (LPIPS)~\cite{zhang2018unreasonable} for visual quality.
For instructional video generation, we further incorporate Fréchet Video Distance (FVD)~\cite{unterthiner2018towards}, VQAScore~\cite{lin2024evaluating}, and Motion Score~\cite{huang2024vbench}. 
VQAScore employs an MLLM to measure image-to-text alignment by computing the probability of a ``Yes" answer to a question: 
\texttt{<Image> Does this figure show \{action instruction\}?}
For Motion Score, we calculate Dynamic Degree and Motion Smoothness as proposed in VBench~\cite{huang2024vbench}, and use their harmonic mean as the final metric. In addition, we utilize ViCLIP~\cite{wang2023internvid} and EgoVLP~\cite{lin2022egocentric} to evaluate text-to-video alignment on SSv2 and Epic100, respectively.



\subsection{Implementation Details}

We implement ShowMe by adapting DynamiCrafter~\cite{xing2025dynamicrafter} through LoRA tuning in two stages. For action-object state manipulation, we first disable the temporal layers and freeze all pretrained parameters, training the LoRA within the projector and spatial transformers for 50K steps using a batch size of 256 and learning rate of 1e-4, followed by 20K steps with structure reward tuning. 
For action-object state prediction, we fine-tune another set of spatiotemporal LoRA for 20K steps using a batch size of 64 with motion consistency reward. Considering the computation cost, we uniformly sample 16 frames for Epic100 and 12 frames for SSv2, resizing to 256$\times$256 resolution during both training and testing phases. During inference, we use the DDIM~\cite{song2020denoising} sampler for 50 denoising steps and apply classifier-free guidance~\cite{ho2022classifier} with a ratio of 7.5 for text conditions. All experiments are conducted using 8 $\times$ RTX 6000 Ada GPUs.

\subsection{Comparisons with State-of-the-Arts Methods}

\begin{figure}[t]
    \centering
    \includegraphics[width=\linewidth]{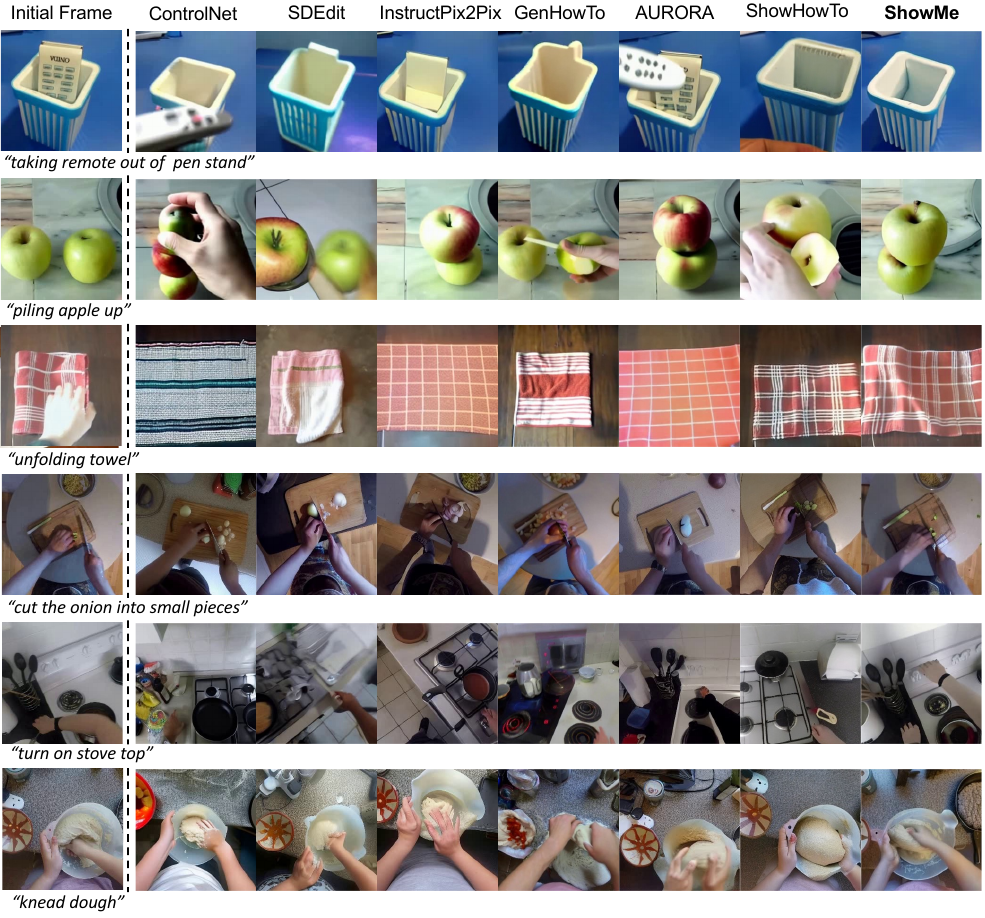}
    \caption{Comparison of different methods for instructional image generation. The first three rows are test samples from SSv2, and the last three rows are from Epic100. Our method is better at completing action instructions and maintaining contextual consistency.}
    \label{fig:ffp}
\end{figure}

\noindent\textbf{Instructional Image Generation.} 
We first compare our method with state-of-the-art image manipulation models, including ControlNet~\cite{zhang2023adding}, SDEdit~\cite{meng2021sdedit}, InstructPix2Pix~\cite{brooks2023instructpix2pix}, GenHowTo~\cite{souvcek2024genhowto}, AURORA~\cite{krojer2025learning} and ShowHowTo~\cite{souvcek2025showhowto}.
As shown in \cref{tab:ffp-sota}, ShowMe consistently outperforms competing methods in image-to-image metrics, achieving the highest CLIP-I and DINO-I scores and the lowest LPIPS on both datasets. This indicates that our model not only preserves semantic similarity but also maintains finer structural details.
Notably, the compared methods are fully fine-tuned on these two datasets, while our model only applies LoRA tuning. 
We also notice that ShowMe has a lower CLIP-T score than all other methods. Interestingly, even ground-truth image-text pairs yield low CLIP-T scores (0.28 on SSv2 and 0.25 on Epic100), suggesting that CLIP-T may struggle to capture alignment between human actions and instructional images. This reveals a crucial limitation in previous methods: while they prioritize text similarity, they often sacrifice contextual fidelity, generating images that match the text superficially but fail to preserve scene consistency. We validate this finding with qualitative examples in \cref{fig:ffp}, showing that while compared models produce semantically relevant images, they often alter key contextual elements, leading to discrepancies in action-object relationships. In contrast, ShowMe effectively balances semantic relevance with contextual consistency.

\begin{figure}[t]
    \centering 
    \includegraphics[width=\linewidth]{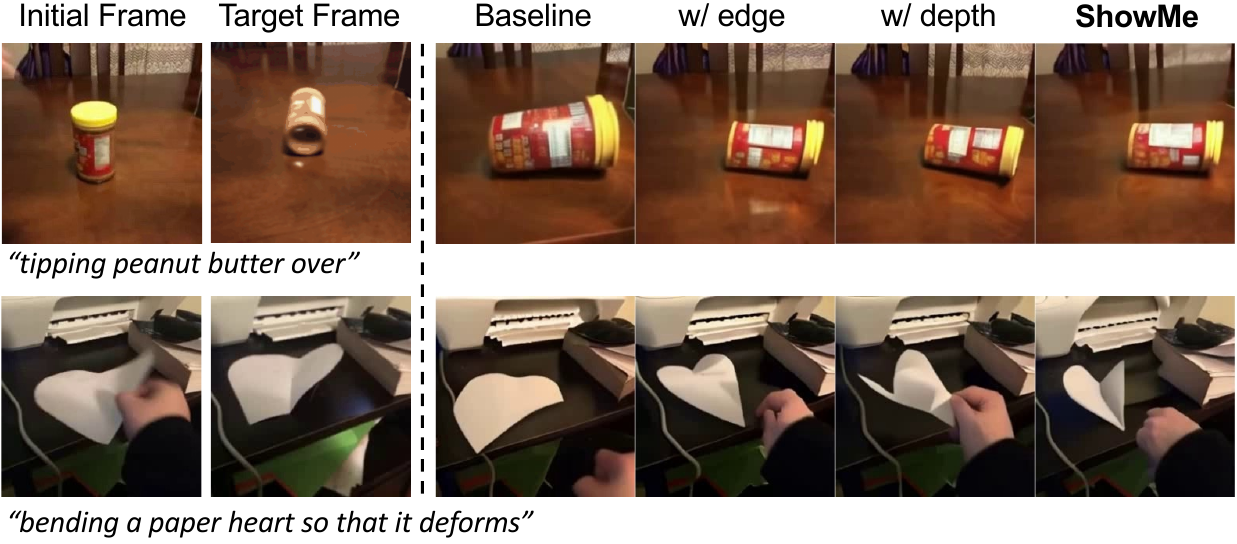}
    \caption{Visualization for the effect of structure reward tuning.}
    \label{fig:ssreward}
\end{figure}

\begin{figure*}[t]
  \centering
   \includegraphics[width=\linewidth]{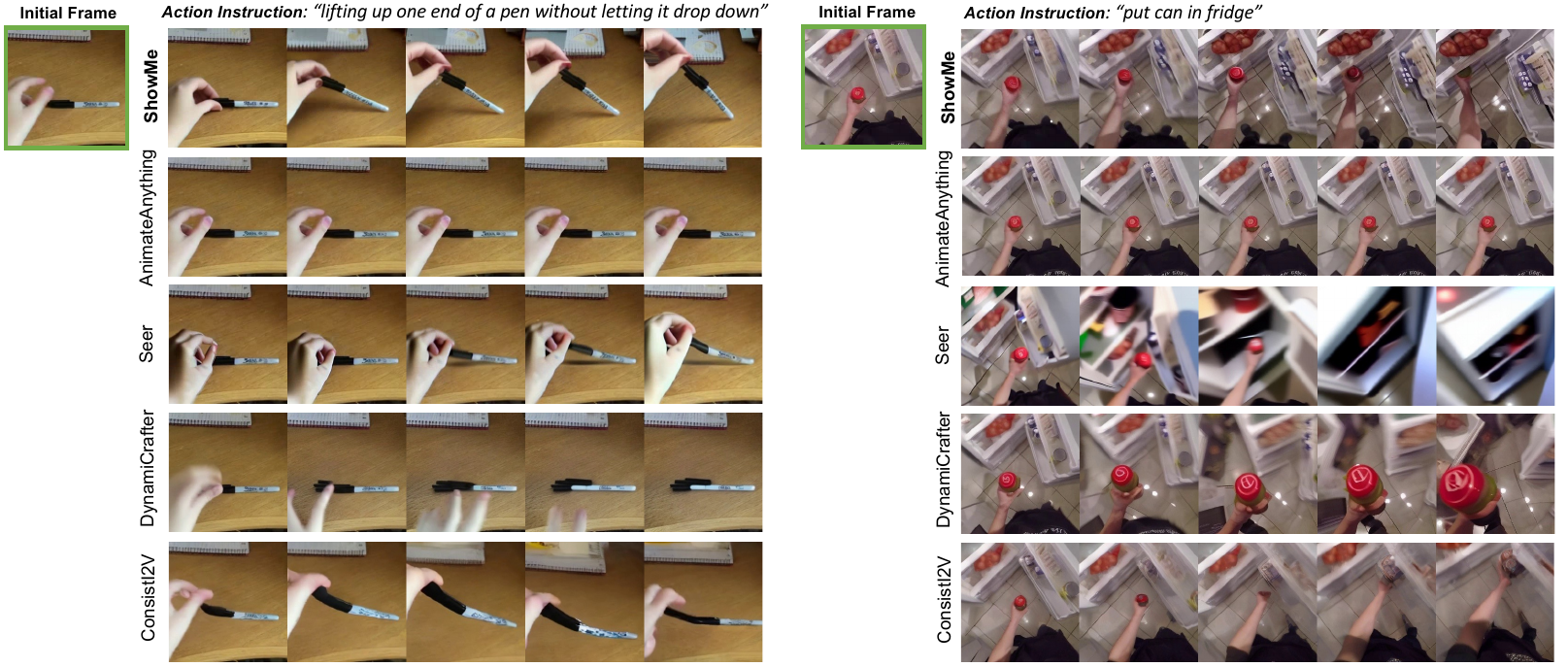}
   \caption{Generated instructional videos from different I2V models for two samples on SSv2 (left) and Epic100 (right).}
   \label{fig:sota-compare}
\end{figure*}

\noindent\textbf{Instructional Video Generation.} We further compare ShowMe with existing state-of-the-art I2V models, including AnimateAnything~\cite{dai2023animateanything}, Seer~\cite{gu2023seer}, ConsistI2V~\cite{ren2024consisti2v}, and DynamiCrafter~\cite{xing2025dynamicrafter}. To ensure a fair comparison, we fully fine-tune all these methods on our curated datasets and report the final results. As shown in \cref{tab:tvp}, ShowMe achieves consistently strong performance across both SSv2 and Epic100, notably delivering the lowest FVD scores on both datasets. While AnimateAnything achieves the highest CLIP-I and the lowest FID, it exhibits the lowest Motion Score, indicating that it tends to generate nearly static frames, as evidenced in \cref{fig:sota-compare}. Compared to Seer and DynamiCrafter, ShowMe maintains strong motion quality while excelling in semantic alignment (CLIP-T and VQA), demonstrating its ability to generate not only plausible movement but also accurate action-object interactions.

\begin{table}[t]
\centering
\renewcommand{\arraystretch}{1.1} 
\resizebox{0.47\textwidth}{!}{ 
\begin{tabular}{lcccccc}
\toprule
Method & CLIP-I $\uparrow$  & CLIP-T $\uparrow$ & DINO-I $\uparrow$ & FID $\downarrow$  & PSNR $\uparrow$ & LPIPS $\downarrow$ \\ 
\hline
baseline & 0.8434 & 0.2827 & 0.5967 & 29.84 & 12.01 & 0.4164 \\ 
w/ $\mathcal{L}_{\text{dp}}$ & 0.8532 & 0.2844 & 0.6255 & 28.25 & 12.56 & 0.3883 \\ 
w/ $\mathcal{L}_{\text{ed}}$ & 0.8553 & 0.2845 & 0.6299 & 26.92 & 12.77 & 0.3806 \\ 
ShowMe & 0.8561 & 0.2860 & 0.6312 & 26.97 & 12.84 & 0.3778 \\ 
\bottomrule
\end{tabular}
}
\caption{Ablation of structure reward tuning on SSv2. `baseline' means LoRA tuning without reward objectives.}
\label{tab:ablation1}
\end{table}

\begin{table}[t]
\centering
\renewcommand{\arraystretch}{1.1} 
\resizebox{0.47\textwidth}{!}{ 
\begin{tabular}{lcccccc}
\toprule
Method & FVD $\downarrow$ & FID$\downarrow$  & CLIP-I $\uparrow$ & ViCLIP $\uparrow$  & VQA $\uparrow$ & Motion $\uparrow$ \\ 
\hline
ShowMe & 74.66 & 10.21 & 0.8584 & 0.2012  & 0.5364   & 94.07 \\ 
\hline
w/o $\mathcal{L}_{\text{mo}}$ & 86.74 & 11.00 & 0.8515 & 0.2003  & 0.5355  & 93.98  \\ 
w/o Stage 1 & 110.73 & 12.54 & 0.8294  & 0.1971  & 0.5237  & 93.73  \\ 
\bottomrule
\end{tabular}
}
\caption{Ablation of core components for instructional video generation on SSv2. `Stage 1' means state manipulation.}
\label{tab:ablation2}
\end{table}


\subsection{Ablation Studies}

\textbf{The effect of structure reward tuning.} We report the ablation results of structure rewards in \cref{tab:ablation1}, showing that each reward contributes uniquely to improving generation quality. $\mathcal{L}_{\text{dp}}$ enhances semantic alignment and perceptual quality, indicating that incorporating depth information helps preserve object structures. $\mathcal{L}_{\text{ed}}$ further refines visual fidelity, leading to sharper details and better structural coherence. When combined, ShowMe achieves the best performance, demonstrating that depth and edge constraints complement each other, resulting in more faithful instructional images.


We further present qualitative results in \cref{fig:ssreward} to highlight the impact of structural rewards. In the first row, the baseline model follows the instruction but fails to capture spatial relationships, resulting in incorrect object sizes and misaligned tag labels. Incorporating edge information preserves texture, while the depth reward improves understanding of rigid structures and 3D spatial layout. When both rewards are combined, the edited image exhibits precise local edits with correct tag positioning and structural integrity.


\begin{figure}[t]
    \centering
    \includegraphics[width=\linewidth]{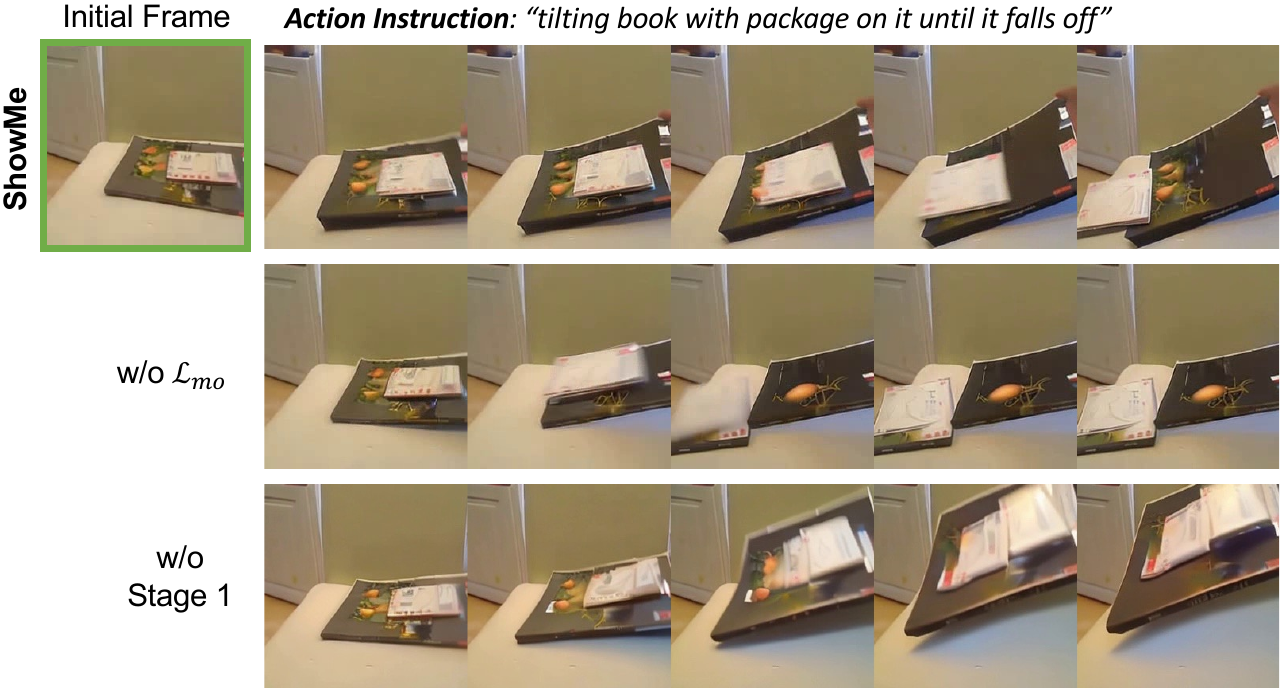}
    \caption{Visualization for the effect of and motion reward tuning and `Stage 1' (action-object state manipulation).}
    \label{fig:motionreward}
\end{figure}

\noindent\textbf{The effect of components for instructional video generation.} 
The ablation results in \cref{tab:ablation2} further highlight the core contributions of ShowMe. Removing the motion consistency reward $\mathcal{L}_{\text{mo}}$ results in degraded temporal coherence, as reflected in increased FVD and reduced CLIP-I and Motion scores. 
Notably, omitting Stage 1 leads to a substantial drop across all metrics, indicating its critical role in ensuring coherent spatial layouts and achieving the intended outcomes. 
Qualitative examples in \cref{fig:motionreward} further illustrate these effects. Without motion rewards, the generated videos exhibit unnatural transitions, such as abrupt object movements during actions like package sliding. When Stage 1 is removed, the model fails to produce the correct final state, \ie, the package does not fall off the book, highlighting a loss in goal-directed reasoning. 
This supports our core claim: instruction-guided state manipulation not only enables meaningful edits but also strengthens video prediction by promoting goal-oriented visual outcomes.



\subsection{Qualitative Analysis}

\noindent\textbf{Comparison of generated instructional videos.} We further present a qualitative comparison with existing I2V generation models, as shown in \cref{fig:sota-compare}. Our observations highlight key differences: (1) AnimateAnything often produces nearly “static” videos, exhibiting minimal human motion and failing to depict dynamic interactions; (2) Seer and ConsistI2V demonstrate some ability to follow instructions but struggle with maintaining visual coherence, either distorting the visual context or failing to preserve the structure of rigid objects; (3) ShowMe, in contrast, exhibits a stronger understanding of instructional prompts and executes them more faithfully. It not only captures clear human-object interactions, but also generates plausible state transformations, resulting in more realistic and contextually accurate video sequences. This highlights ShowMe’s advantage in both semantic comprehension and visual consistency.

\begin{figure}[t]
    \centering
    \begin{subfigure}[b]{\linewidth}
        \centering
        \includegraphics[width=\linewidth]{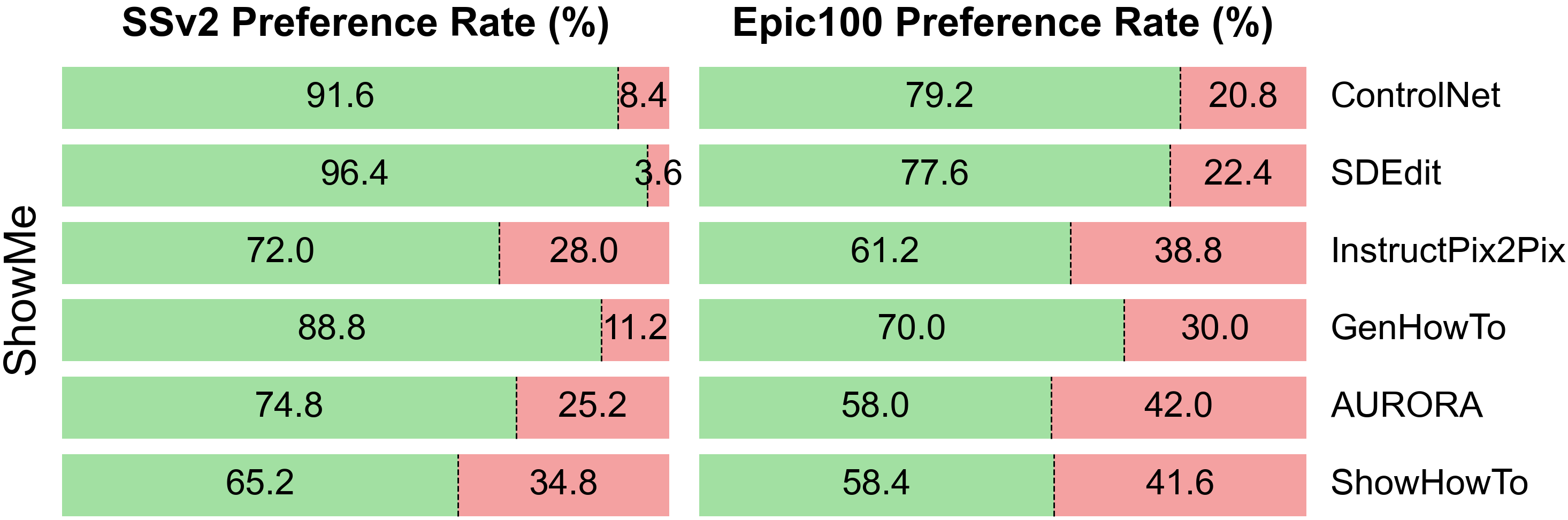}
        \caption{Instructional image generation.}
        \label{fig:user_study_qual}
    \end{subfigure}
    \vspace{-2mm}
    
    \begin{subfigure}[b]{\linewidth}
        \centering
        \includegraphics[width=\linewidth]{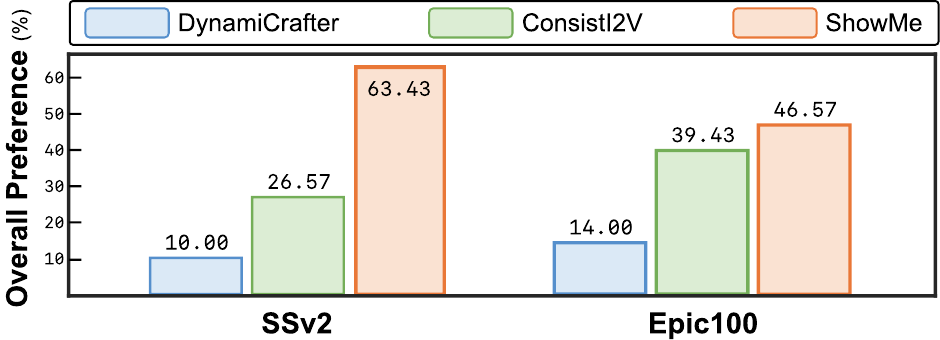}
        \caption{Instructional Video generation.}
        \label{fig:user_study_quant}
    \end{subfigure}
    
    \caption{Human evaluation results. We conduct pairwise comparisons for instructional image generation and adopt a ``winner-takes-all" setting for video generation.
    }
    \label{fig:user_study}
\end{figure}
\noindent\textbf{Human Evaluation.} 
Finally, we conduct a user study to evaluate the advantages of our method. We sample a subset of 50 examples for each dataset and collect the generated results from all baseline models. Each sample is evaluated by five Amazon Mechanical Turk workers, who are asked to select the candidate that best executes the instructed action while preserving the context of the original image.  As shown in Figure~\ref{fig:user_study_qual},
ShowMe outperforms other models by a great margin on SSv2, achieving over 90\% win rates against ControlNet and SDEdit, demonstrating its strengths in contextual consistency and non-rigid editing.
Notably, while ShowMe consistently outperforms other baselines on Epic100, the preference gap is smaller than that in SSv2. We argue that SSv2 involves diverse, nuanced daily-life actions requiring deeper reasoning, whereas Epic100’s constrained kitchen scenarios present fewer challenges. This can be further supported by the video preference rate in Figure~\ref{fig:user_study_quant}, where ShowMe outperforms ConsistI2V by 36.86\% on SSv2 and only 7.14\% on Epic100.

\noindent\textbf{Limitations.} 
While our model can manipulate action object states, it struggles with complex edits from short instructions and with associating actions to non-salient objects correctly, sometimes producing objects “out of thin air” rather than manipulating existing ones. In addition, despite using structure reward tuning, the model may still struggle to distinguish the foreground in dynamic scenes, leading to object distortions and artifacts. Finally, our method currently generates only short clips rather than full videos, so extending it to long-term instructional tasks is a promising direction.



\section{Conclusion}


In this paper, we introduce ShowMe, a unified framework that reimagines diffusion models for both instructional image and video generation. We highlight the limitations of treating these tasks in isolation: image manipulation lacks temporal context, while video prediction often overlooks the intended outcome. ShowMe addresses these gaps by selectively activating spatial and temporal components and employing a two-stage tuning strategy with structure and motion rewards. 
Experiments show that ShowMe surpasses strong baselines, underscoring the potential of VDMs as a holistic solution for visual instruction generation.



\section*{Acknowledgements}
This work was partially supported by the Office of Naval Research (ONR) grant (N00014-23-1-2417), and the Army Research Office (ARO) grant (W911NF-24-1-0385). Any opinions, findings, and conclusions or recommendations expressed in this material are those of the authors and do not necessarily reflect the views of ONR or ARO.
{
    \small
    \bibliographystyle{ieeenat_fullname}
    \bibliography{main}
}

\clearpage
\setcounter{page}{1}
\setcounter{section}{0}
\maketitlesupplementary

\renewcommand{\thesection}{\Alph{section}}


\section{Data Curation}

\subsection{Something-Something V2}

The Something-Something v2 (SSv2) dataset is a comprehensive collection of labeled video clips illustrating human hand gestures and interactions with objects. It contains a diverse range of actions such as ``Putting something into something," ``Turning something," and ``Pushing something from left to right." The original dataset comprises approximately 220,847 video clips, each around 4 seconds long, spanning 174 action categories. Compared to Epic100, SSv2 videos clearly exhibit an initial state and generally conclude with a frame depicting action completion. However, we identified several action categories without distinct human actions or meaningful object state changes, such as ``holding," ``showing," ``fail," and ``nothing happens," along with potentially harmful actions like ``hitting" and ``throwing." To ensure dataset relevance and safety, we manually reviewed all 174 action templates, identified and filtered out 62 categories containing keywords unlikely to reflect meaningful or useful changes. As a result, we curated the dataset to include 112,321 training samples and 2,048 test samples.

\subsection{EPIC-KITCHENS 100}
\begin{figure}[t]
    \centering
    \includegraphics[width=\linewidth]{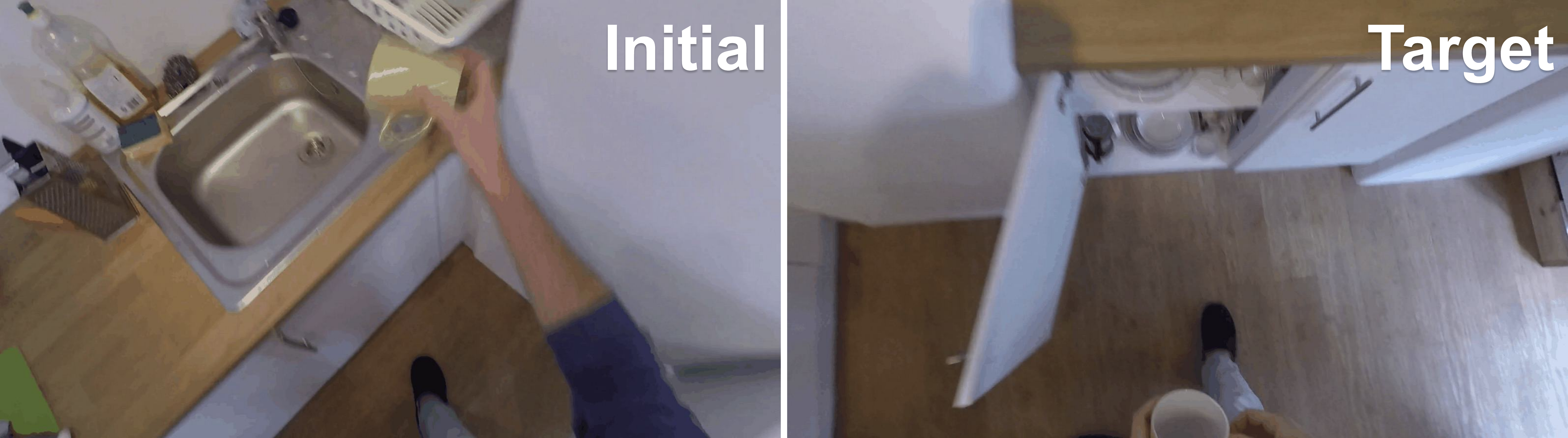}
    \vspace{-0.5cm}
    \caption{Inconsistent visual context in egocentric instructional videos. From left to right: initial and target frames.}
    \label{fig:context}
    \vspace{-0.2cm}
\end{figure}

The EPIC-KITCHENS (Epic100) dataset comprises approximately 90,000 egocentric action segments with 20,000 unique narrations, covering 97 verb classes and 300 noun classes, recorded from daily kitchen activities. However, the continuous and unscripted nature of the recordings leads to challenges in generating coherent instructional videos. Specifically, egocentric videos often include highly dynamic actions and abrupt scene transitions, as shown in \cref{fig:context}, and the end frames of narrations may not precisely align with the corresponding video clips. To address these challenges, we introduce a dual-similarity filtering approach to identify and remove videos with significant scene changes or misaligned content.

Our dual-similarity strategy involves two metrics:
\begin{itemize}
    \item \textbf{Semantic Similarity}: We compute the similarity between text narrations and candidate frames using CLIP embeddings to ensure semantic alignment.
    \item \textbf{Visual Consistency}: We compute a framewise DINO similarity score between the initial frame and each of the downsampled last 8 frames.
\end{itemize}

By multiplying these two similarity scores, we measure the in-context semantic relevance and visual coherence of target frames. The frame with the highest combined similarity score is selected as the candidate target frame if its score exceeds a predefined threshold (We empirically set it to 0.1). Using this filtering strategy, we refined the dataset to 57,602 training samples and 8,236 test samples.


\section{Ego4D Evaluation}

To validate the effectiveness of our method, we followed LEGO’s benchmark~\cite{lai2024lego} setting and conducted additional experiments on Ego4D for instructional image generation. As shown in Table~\ref{tab:ego4d-img}, our method achieves competitive results compared to LEGO. Importantly, LEGO fully fine-tunes Stable Diffusion with MLLM-enriched text instructions, whereas we only apply LoRA tuning with the vanilla action instructions provided in Ego4D. We further extend it for instructional video generation,  where ShowMe consistently improves over the baseline and achieves the best FVD among all compared methods, as illustrated in Table~\ref{tab:ego4d-vid}.

\setlength{\intextsep}{5pt}
\begin{table}[h]
\centering
\resizebox{0.49\textwidth}{!}{
\begin{tabular}{lcccccccc}
\toprule
Methods & EgoVLP $\uparrow$  & EgoVLP$^{+}$ $\uparrow$ & CLIP-I $\uparrow$ & FID $\downarrow$ & PSNR $\uparrow$ & LPIPS $\downarrow$ & BLIP-B $\uparrow$ & BLIP-L $\uparrow$ \\
\midrule
ProxEdit & 44.51 & 72.68 & 68.17 & 33.01 & 11.88 & 40.90 & 17.73 & 17.35 \\
SDEdit & 50.07 & 72.90 & 73.35 & 33.35 & 11.81 & 41.60 & 19.80 & 19.74\\
IP2P & 62.19 & 78.84 & 78.75 & 24.73 & \underline{12.16} & 37.16 & \underline{20.00} & 20.56 \\
LEGO & \underline{65.65} & \textbf{80.44} & \textbf{80.61} & 23.83 & \textbf{12.29} & \underline{36.43} & \textbf{20.38} & \underline{20.70} \\
\textbf{ShowMe} & \textbf{66.93} & \underline{79.14}  & \underline{79.77}  & \textbf{19.38}  & 11.92 & \textbf{35.59}  & 19.76  & \textbf{22.28} \\
\bottomrule
\end{tabular}
}
\caption{Comparison of instructional image generation on Ego4D.}
\label{tab:ego4d-img}
\end{table}

\begin{table}[h]
\centering
\resizebox{0.49\textwidth}{!}{
\begin{tabular}{lcccccccc}
\toprule
Methods & FVD $\downarrow$ & FID $\downarrow$ & CLIP-I $\uparrow$  & CLIP-T $\uparrow$  & EgoVLP $\uparrow$ & Motion $\uparrow$ \\
\midrule
AnimateAnything & 271.14 & 13.24 & \textbf{0.8805} & 0.2541 & 0.3123 & 14.80 \\
ConsistI2V & \underline{81.60} & \textbf{8.72} & \underline{0.8768} & 0.2585 & \textbf{0.3628} & 74.61 \\
DynamiCrafter & 105.76 & 16.20 & 0.8401 & \underline{0.2601} & 0.3064 & \underline{96.69} \\
\textbf{ShowMe} & \textbf{72.51} & \underline{12.00}  & 0.8609  & \textbf{0.2613}  & \underline{0.3196} & \textbf{96.75}  \\
\bottomrule
\end{tabular}
}
\caption{Comparison of instructional video generation on Ego4D.}
\label{tab:ego4d-vid}
\end{table}

\section{More Implementation Details}
\begin{figure}[t] 
    \centering 
    \includegraphics[width=\linewidth]{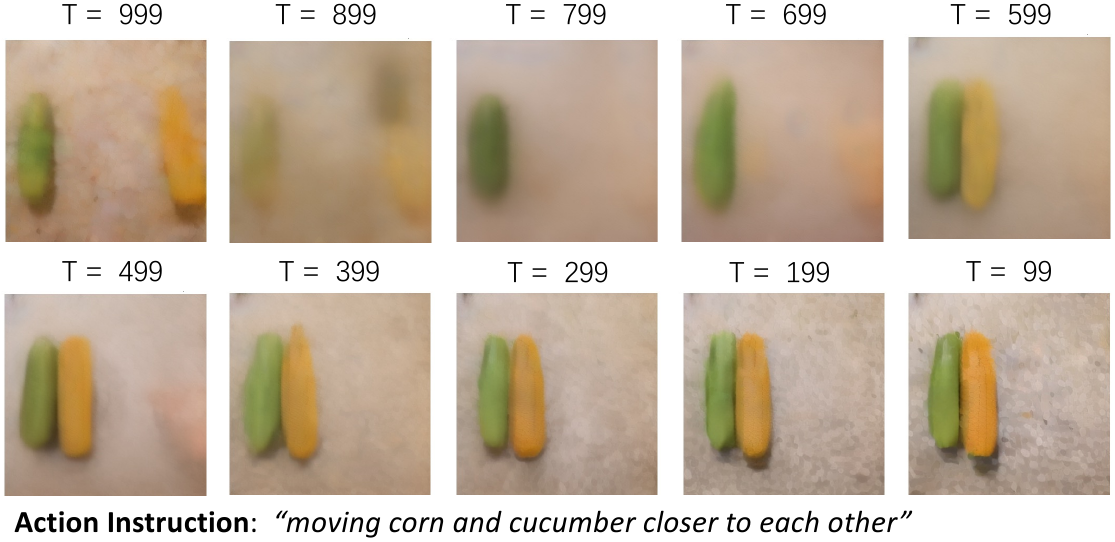}
    \caption{Visualization of one-step denoising results at different diffusion steps.}
    \label{fig:rationale}
\end{figure}

For action-object state manipulation and prediction, we set the LoRA rank to 128 and 64, respectively, with a dropout rate of 0.1. For structure reward tuning, we empirically set the diffusion threshold $\Gamma$ for the one-step approximation to 200, and to 500 for motion reward tuning, as evidenced in~\cref{fig:rationale}.
The weights for $\mathcal{L}_{\text{dp}}$ and $\mathcal{L}_{\text{ed}}$ are both set to 1, while the weight for $\mathcal{L}_{\text{mo}}$ is set to 0.001. During reward tuning, these loss functions work together with the noise prediction loss to optimize the model.

\section{Reward Computation Cost}
Table~\ref{tab:reward_complexity} shows that the dominant cost in Stage-1 is the grad-enabled VAE decode, while the rewards themselves add only modest increments. In Stage-2, the motion reward operates entirely in latent space, incurring only a moderate computational cost. In practice, the observed training throughput reduction is acceptable given the gains in alignment and motion quality.

\begin{table}[h]
\centering
\resizebox{0.48\textwidth}{!}{%
\begin{tabular}{llccc}
\toprule
Stage & Component & $\Delta$VRAM Peak & Time (ms) & $\Delta$GFLOPs \\
\midrule
\multirow{3}{*}{Stage 1}
 & VAE decode (grad)        & +1.92 GB & 187  & --   \\
 & + Edge reward  & +0.19 GB & 368  & 810  \\
 & + Depth reward  & +0.14 GB & 1022 & 641  \\
\midrule
Stage 2
 & + Motion reward          & +0.79 GB & 453  & 1559 \\
\bottomrule
\end{tabular}
}
\caption{Complexity of reward tuning (per sample at $256^2$, fp16). 
}
\label{tab:reward_complexity}
\end{table}

\section{Rationale for One-Step Approximation}
We visualize one-step denoising samples at different diffusion steps during training, as shown in~\cref{fig:rationale}. It can be observed that at lower noise levels, the approximation closely resembles the target image and preserves spatial details well. In contrast, at higher noise levels (above 600), although structural information is lost, the approximated samples still capture the overall motion trends. This behavior aligns with the sampling process during inference: the model typically plans the global layout and motion patterns in the early high-noise steps, while refining spatial and structural details in the later denoising stages.






\begin{figure}[h] 
    \centering 
    \includegraphics[width=\linewidth]{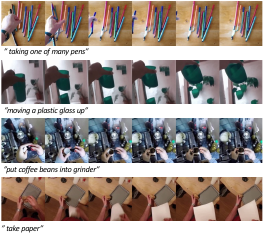}
    \caption{Failure examples. The first two rows are from SSv2 and the last two rows are from Epic100.}
    \label{fig:failure}
\end{figure}

\section{Limitations and Future Work}
Finally, we present several failure cases in Figure~\ref{fig:failure} to illustrate the limitations of our method. In the first and third examples, when a scene contains multiple similar or complex objects, the model struggles to ground the instruction precisely. For instance, distinguishing between ``coffee beans" and the ``grinder", which leads to task failure. In the second example, although the action is correct, the model’s limited understanding of 3D space causes the plastic cup to appear split in half. In the final example, the model hallucinates a blank sheet of paper, indicating a failure to recognize the paper already being held. One possible solution is to introduce spatial grounding models for accurate target localization. In addition, explicitly conditioning on segmentation and depth information, along with using MLLMs to generate more detailed instructions, could further improve performance.




\begin{figure*}[t]
    \centering
    \includegraphics[width=\linewidth]{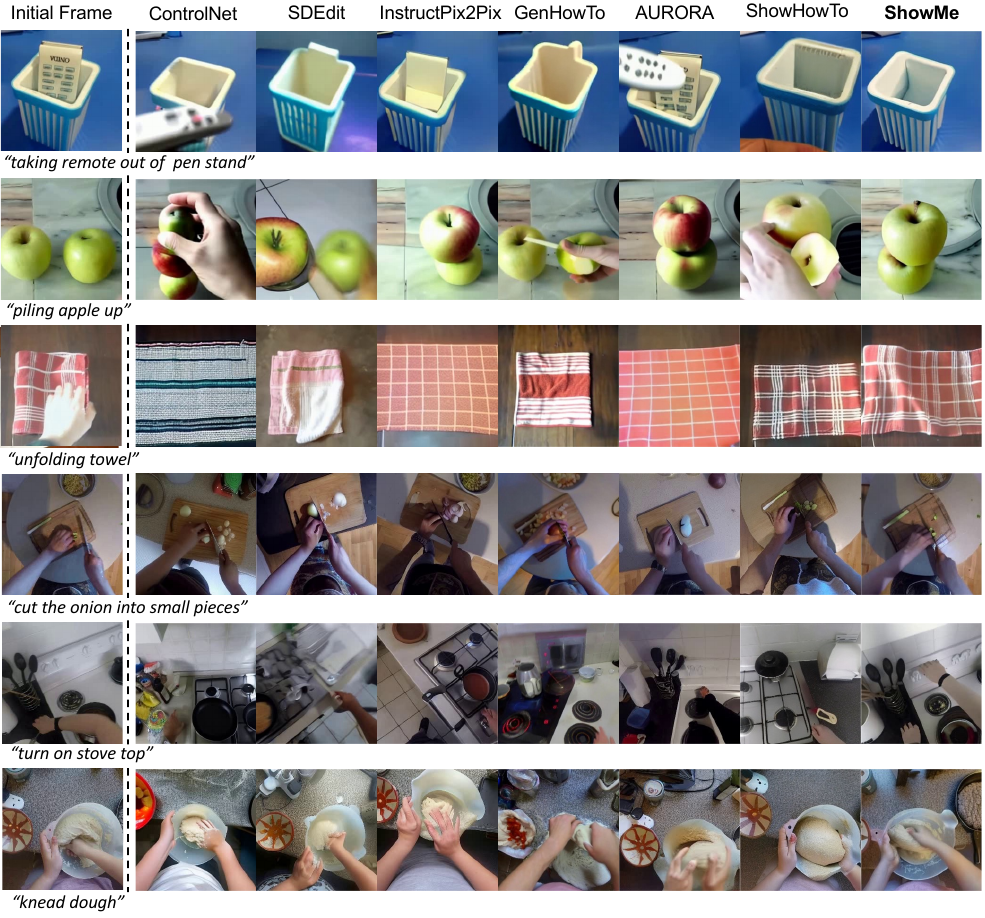}
     \caption{Comparison of different methods for instructional image generation. The first three rows are test samples from SSv2, and the last three rows are from Epic100. Our method is better at completing action instructions and maintaining contextual consistency.}
    \label{fig:ffp-large}
\end{figure*}

\begin{figure*}[t]
    \centering
    \includegraphics[width=\linewidth]{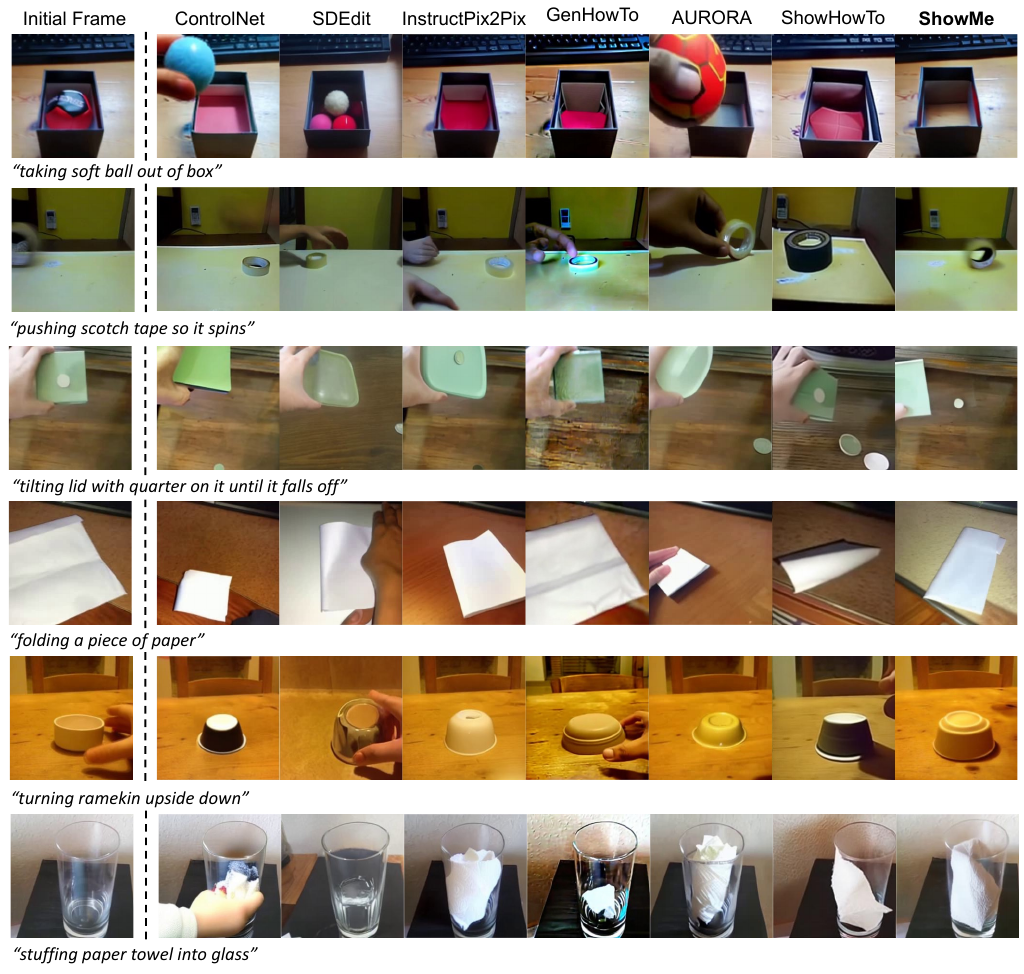}
    \caption{More visualization results of instructional image generation on SSv2.}
    \label{fig:ffp2}
\end{figure*}

\begin{figure*}[t]
    \centering
    \includegraphics[width=0.9\linewidth]{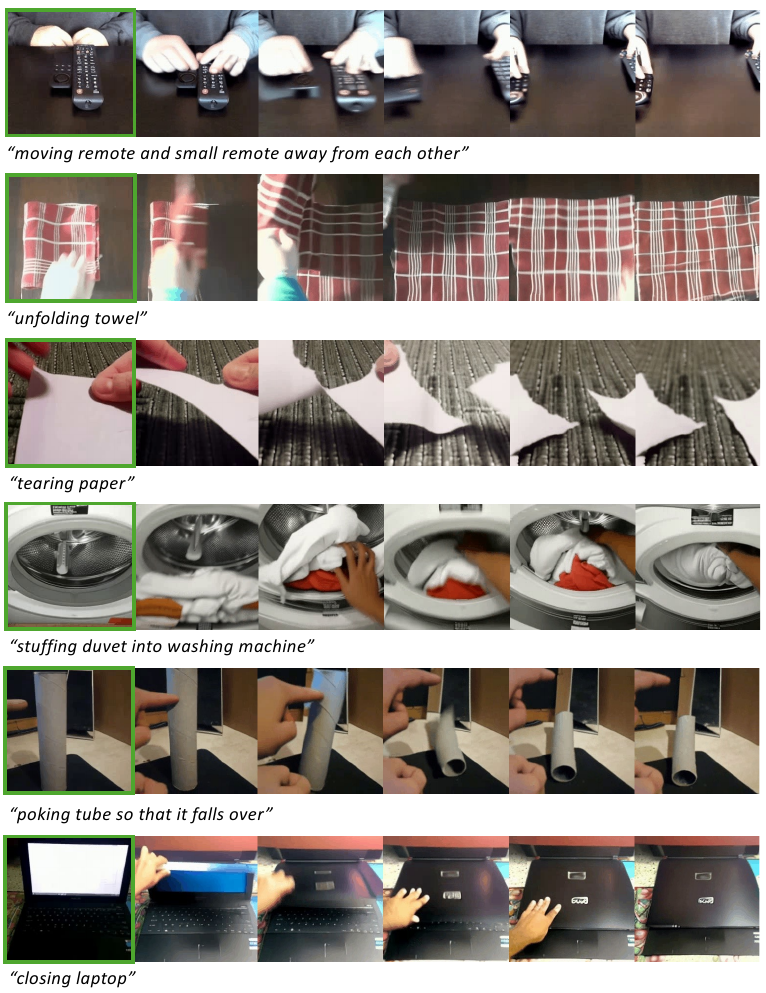}
    \caption{Generated videos on SSv2 dataset.}
    \label{fig:ssv2-vis}
\end{figure*}

\begin{figure*}[t]
    \centering
    \includegraphics[width=0.9\linewidth]{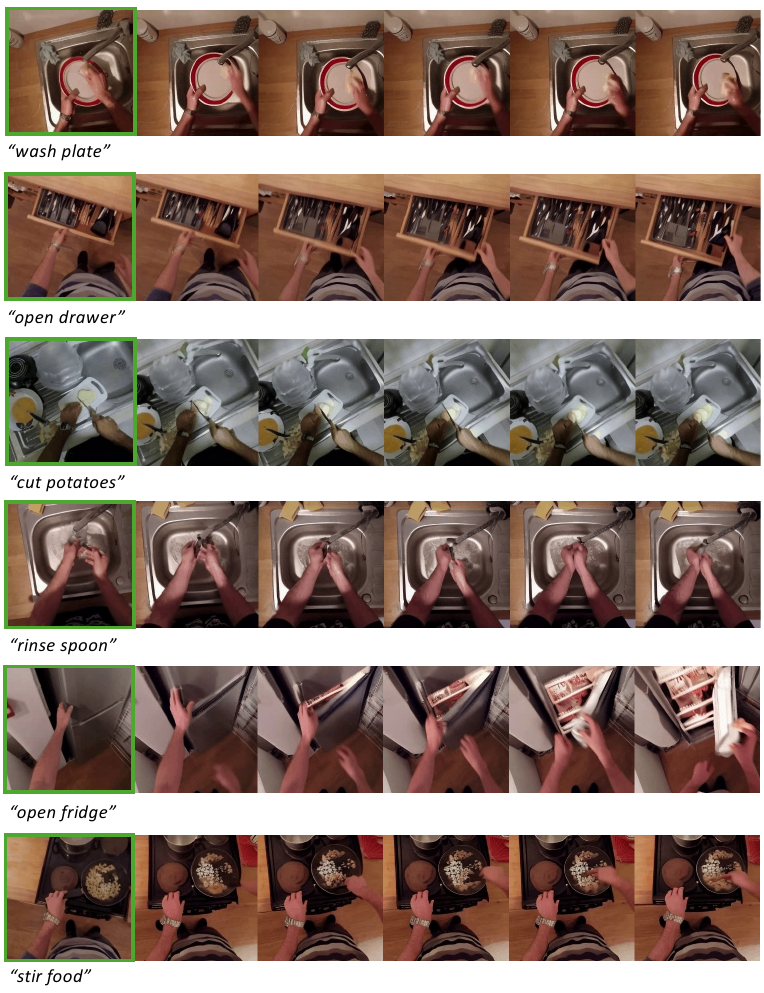}
    \caption{Generated videos on Epic100 dataset.}
    \label{fig:epic-vis}
\end{figure*}

\end{document}